\documentclass[12pt]{article}

\usepackage{amsmath}
\usepackage{newtxtext,newtxmath}
\usepackage[letterpaper,margin=1in]{geometry}
\usepackage{graphicx}
\usepackage{float}
\date{}

\renewenvironment{abstract}
    {\quotation}
    {\endquotation}

\makeatletter
\renewcommand{\fnum@figure}{\textbf{Figure \thefigure}}
\renewcommand{\fnum@table}{\textbf{Table \thetable}}
\makeatother

\usepackage{scicite}
\usepackage{url}

\usepackage{booktabs}
\usepackage{tabularx}
\usepackage{array}
\usepackage{ragged2e}
\usepackage[table]{xcolor}
\usepackage{makecell}
\usepackage{tcolorbox}
\usepackage{needspace}

\newcolumntype{Y}{>{\RaggedRight\arraybackslash}X}

\newcounter{conceptbox}
\newenvironment{conceptbox}[2]{%
    \refstepcounter{conceptbox}%
    \begin{tcolorbox}[
        float=t!,
        width=0.94\linewidth,
        center,
        colback=black!3,
        colframe=black!45,
        boxrule=0.55pt,
        arc=0pt,
        outer arc=0pt,
        boxsep=0pt,
        left=7pt,
        right=7pt,
        top=5pt,
        bottom=5pt,
        before skip=0pt,
        after skip=0pt
    ]%
    \linespread{1.05}\small
    \setlength{\parindent}{0pt}%
    \setlength{\parskip}{3pt}%
    \textbf{Box~\theconceptbox: #1}\label{#2}\par
}{%
    \end{tcolorbox}%
}

\def\scititle{Toward Trustworthy Physical AI for Human Interaction}

\title{\bfseries \boldmath \scititle}

\author{
    Niccol\`o~Pagliarani$^{1\dagger}$,
    Maximilian~St\"olzle$^{2\dagger}$,
    \and
    Cosima~du~Pasquier$^{3}$,
    Jeff~Lui$^{4}$,
    Andrew~Sabelhaus$^{5}$,
    \and
    Cecilia~Laschi$^{6}$,
    Gentiane~Venture$^{4}$,
    Daniela~Rus$^{2\ast}$,
    Matteo~Cianchetti$^{1\ast}$
    \and
    \small$^{1}$The BioRobotics Institute, Scuola Superiore Sant'Anna, Pontedera, Italy.
    \and
    \small$^{2}$CSAIL, Massachusetts Institute of Technology, Cambridge, MA, USA.
    \and
    \small$^{3}$CHARM Lab, Stanford University, Stanford, CA, USA.
    \and
    \small$^{4}$The University of Tokyo, Tokyo, Japan.
    \and
    \small$^{5}$Boston University, Boston, MA, USA.
    \and
    \small$^{6}$Department of Mechanical Engineering, College of Design and Engineering,\\
    \small and Advanced Robotics Centre, National University of Singapore, Singapore.
    \and
    \small$^\ast$Corresponding authors: Daniela Rus and Matteo Cianchetti.
    \and
    \small$^\dagger$These authors contributed equally to this work.
}

\begin{document}

\maketitle

\begin{abstract} \bfseries \boldmath
Robots are entering human spaces faster than we can establish when they deserve trust. We propose a framework for trustworthy physical AI that integrates Safety, Behavioral Intelligibility, and Perceptual Alignment across embodiment, control, cognition, and design. Trustworthiness emerges from aligning physical capabilities, observable behavior, and expectations people form during interaction.

\end{abstract}

\noindent
Robotics has long promised machines that can leave structured industrial settings and operate in the spaces where people live, work, heal, and receive care~\cite{desantis2008atlas,he2021challenges}. Yet this transition has been slower than the field often suggests. Human environments demand robots that can perform useful tasks while remaining safe to interact with, understandable in their behavior, and clear about their capabilities~\cite{haddadin2013towards,khan2014compliance,rubagotti2022perceived,kok2020trust}. These qualities are closely connected: the way a robot is built shapes how it behaves, and both influence what people expect from it.  We use \textit{trustworthiness} to capture how these qualities come together in supporting appropriate human reliance. A robot designed to limit contact loads may still be difficult to anticipate, while predictable behavior can coexist with harmful contact. Appearance also matters, as an approachable form can shape expectations before the robot has demonstrated what it can do~\cite{dragan2013legibility,zlotowski2016appearance,goetz2003matching,paepcke2010judging}. This reveals a broader challenge, as properties usually studied separately come together in the same physical interaction. Understanding how they influence one another motivates our framework for trustworthy physical AI~\cite{he2021challenges,devisser2020towards}. The challenge is becoming commercially urgent as domestic and humanoid robots move from research laboratories into homes and workplaces. More than 50 million Roombas have already operated in private homes~\cite{robotreport2025look}, while companies including Tesla, Figure, Sunday Robotics, and 1X are developing platforms explicitly intended for human-centered environments.
In this context, \textit{trust} and \textit{trustworthiness} should not be conflated. Trust is a human-side relational variable, shaped by uncertainty, prior experience, context, and perceived reliability. Trustworthiness, by contrast, is a context-dependent system property: the extent to which a robot’s embodiment, behavior, and communicative cues support appropriate reliance by both direct users and bystanders during physical interaction~\cite{lee2004trust,kok2020trust,firminodesouza2025trust}. The design objective is therefore to maximize the robot’s trustworthiness across the intended interaction contexts, including for people with limited familiarity over their proximity to it.

\begin{figure}[H]
    \centering
    \includegraphics[width=\textwidth]{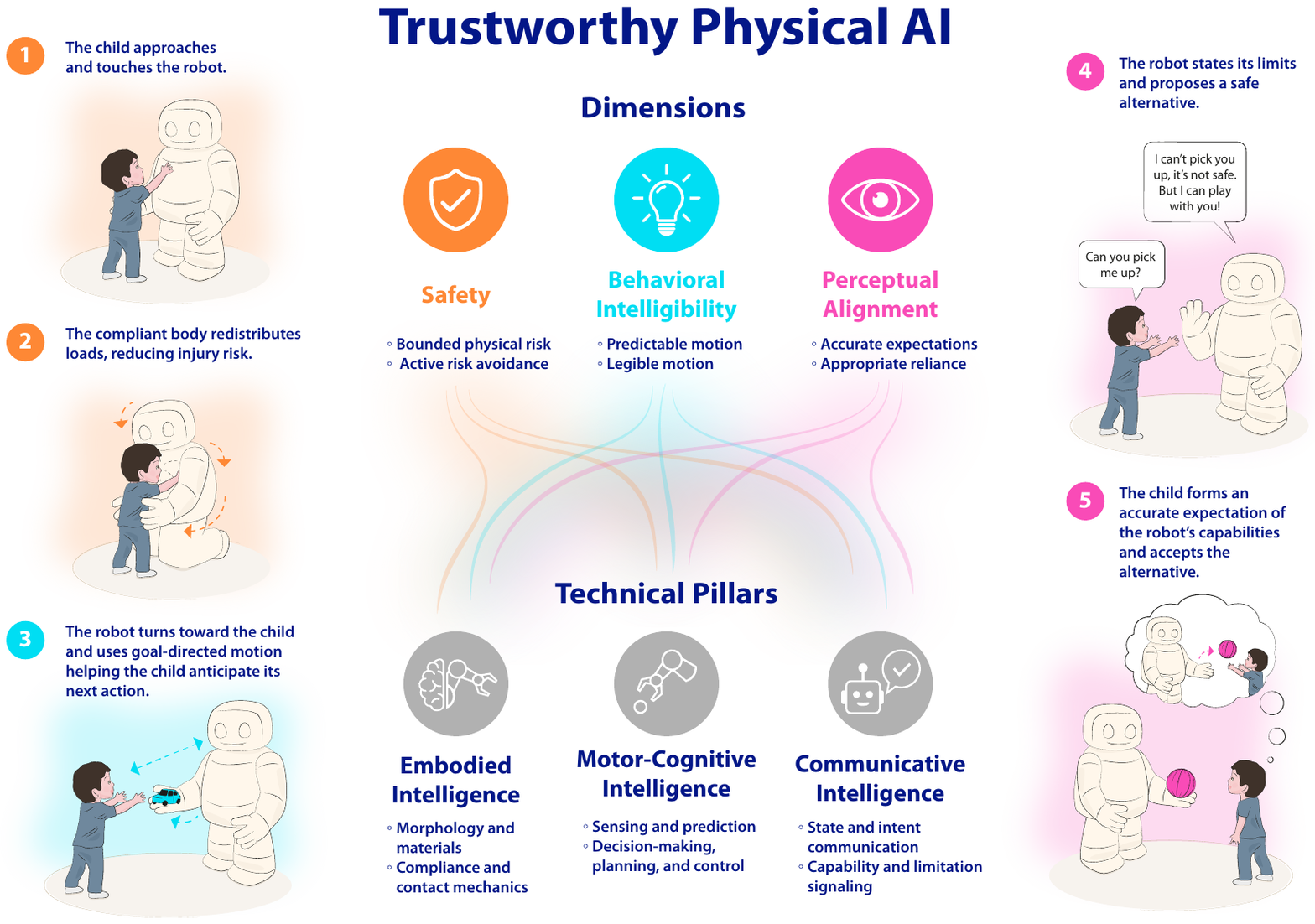}
    \caption{\textbf{Framework for trustworthy physical AI.}
    Trustworthy physical AI emerges from the interplay between three system-side dimensions---\textbf{Safety}, \textbf{Behavioral Intelligibility}, and \textbf{Perceptual Alignment}---and three technical pillars---\textbf{Embodied Intelligence}, \textbf{Motor-Cognitive Intelligence}, and \textbf{Communicative Intelligence}. The interaction scenarios illustrate how these pillars jointly bound physical risk, make robot behavior anticipatable, and enable users to form accurate expectations of the robot's capabilities and limitations.}
    \label{fig:trust_framework}
\end{figure}

We propose an integrated framework for \textbf{trustworthy physical AI} that connects the qualities required for trustworthy interaction with the technical means through which they can be achieved, as illustrated in Fig.~\ref{fig:trust_framework}. We identify three \textbf{system-side dimensions} that we regard as necessary for trustworthy physical interaction: \textbf{Safety}, \textbf{Behavioral Intelligibility}, and \textbf{Perceptual Alignment}. These dimensions address demands that become especially salient as robots enter human environments involving non-expert users and bystanders.

\textbf{Safety} requires physical risk to remain within context-specific bounds and the robot to recover safely when nominal operation is disrupted~\cite{luckcuck2019formal,belta2019formal}. \textbf{Behavioral Intelligibility} captures whether people can anticipate the robot's actions and infer its goals. \textbf{Perceptual Alignment} concerns whether interaction supports an accurate understanding of the robot's capabilities and role. The meaning and relative importance of these dimensions depend on the interaction context. Broader concerns, including accountability, remain essential and complement the physical-interaction focus developed here.

Three \textbf{technical pillars} provide complementary means of shaping these dimensions. \textbf{Embodied Intelligence} captures how the robot's physical structure and dynamics mediate interaction~\cite{mengaldo2022concise}. \textbf{Motor-Cognitive Intelligence} connects sensing and decision making to controlled physical action. \textbf{Communicative Intelligence} concerns how the robot makes its state, intentions, and role available to the people interacting with it.

Each dimension draws on multiple pillars, as detailed in Table~\ref{tab:trustworthiness_matrix}. 
The dimensions define the desired outcomes of trustworthy interaction, while the pillars identify the means through which they can be pursued.

Our position is that \textbf{claims of trustworthiness should be supported by evidence across all three dimensions within the intended interaction context}: bounded physical risk, behavior that people can anticipate, and expectations consistent with demonstrated capabilities and limitations. Their alignment should therefore guide design and evaluation from the outset. Achieving this alignment requires jointly considering how embodiment, control, cognition, and communicative cues shape physical interaction and human reliance. Our framework provides a common structure for examining these dependencies and translating them into actionable design recommendations and candidate evaluation criteria. These establish a research agenda for developing and validating the proposed requirements across users, tasks, and operating conditions.

\section*{Safety}
Safety is the capacity to keep physical risk within a task- and user-specific envelope before, during, and after contact. Following established machinery-safety terminology, harm refers to physical injury or damage to health, and risk combines the severity of possible harm with the probability of its occurrence~\cite{iso2010iso12100}.
Many interactive tasks require proximity or deliberate contact~\cite{desantis2008atlas,lasota2017survey}, which means that robots should prevent unwanted or inappropriate contact when possible and bound the consequences of intended or unavoidable contact.
Separation, collision avoidance, and speed limits remain essential in industrial collaborative operation~\cite{iso2016ts15066}, but applied too conservatively, they can preclude useful physical interaction~\cite{lasota2017survey}.

Failure can arise on either side of this trade-off. Avoidance-only systems may halt or withdraw so often that meaningful collaboration becomes impossible~\cite{lasota2017survey}. 
Conversely, active safety may fail under loss of observability, inaccurate human-motion prediction, model mismatch, delayed feedback, or infeasible real-time safety filtering~\cite{laine2020eyes,fisac2018probabilistically,bajcsy2021robust,hsu2024safety}. If contact occurs, even a compliant-looking robot may generate harmful loads through high speed, reflected inertia, concentrated geometry, stored elastic energy, gravity, or aggressive feedback\cite{haddadin2013towards,mansfeld2018safety,dickson2025safe}. Risk may also increase when ambiguous modes or poorly communicated hazards encourage unsafe proximity, hand placement, workspace entry, or timing~\cite{gopinath2019understanding}. Safety must therefore address both the likelihood and severity of harm, including failure detection and recovery.

When contact occurs, embodied and motor-cognitive intelligence jointly shape the magnitude, rate, and duration of mechanical loading~\cite{desantis2008atlas,haddadin2013towards}, while severity also depends on contact geometry, the exposed body region, and the person's capacity to recoil\cite{behrens2022statistical,mansfeld2018safety}. Acceptable limits depend on the user and context\cite{samarathunga2025assessing}; interaction near a premature infant's head or airway, for example, requires evidence specific to that vulnerability. Safety claims should therefore specify intended users, task, environment, permitted contact modes, and assumptions about training or protective equipment.

Evaluation must first delimit the interaction envelope and identify relevant hazards~\cite{iso2010iso12100}. To assess the potential severity of harm, the formalization below focuses on mechanical loading during blunt impact and sustained contact. Clamping, sharp or tool-mediated contact, and thermal, electrical, or chemical hazards require additional criteria~\cite{haddadin2013towards,mansfeld2018safety,samarathunga2025assessing}.
Industrial standards specify force-, pressure-, and body-region-specific limits~\cite{iso2016ts15066}; biomechanics also studies deformation, loading velocity, transferred energy, and acceleration~\cite{haddadin2013towards,mansfeld2018safety}. Standardized methods measure and analyze contact force and pressure~\cite{iso2023pas5672}. The associated human-tolerance limits are commonly estimated from controlled impacts or quasi-static compression tests~\cite{behrens2022statistical}, providing reference bounds under those test conditions. Predicting the load generated by a particular robot requires a closed-loop contact model accounting for the robot’s mechanical properties, contact evolution, and controller dynamics.

To formalize this coupling, let \(u=\pi_\theta(q,\dot q,z)\) be the policy output and let \(\chi_\mathrm{c}=(\chi_\mathrm{R},\chi_\mathrm{H},z_\mathrm{c})\) collect the contact parameters.
The full-order contact dynamics can then be written as
\begingroup
\small
\begin{equation}
    M(q)\ddot q+h(q,\dot q)=B(q)u+\tau_\mathrm{c},
    \quad
    \tau_\mathrm{c}=-\int_{\mathcal{A}_\mathrm{c}(t)}J_r^\top(q,r) n(r,q)\,p(t,r)\,\mathrm{d}A,
    \quad
    0=\Psi_\mathrm{c}\!\left(q,\dot q,\ddot q,p;\chi_\mathrm{c}\right).
    \label{eq:full_order_contact_safety}
\end{equation}
\endgroup
Here \(q\) is the robot configuration, \(M(q)\) is its configuration-dependent inertia matrix, \(h\) collects Coriolis, gravitational, elastic, damping, and other bias forces, and \(B(q)\) maps the policy output \(u\) to generalized forces; \(z\) collects task, perception, and environment context. The Jacobian \(J_r(q,r)\) maps configuration velocity to the velocity of surface point \(r\), \(n(r,q)\) is the local unit surface normal, and \(p(t,r)\) is the local contact pressure; their surface integral produces the generalized contact force \(\tau_\mathrm{c}\)~\cite{khatib1987unified}. The relation \(\Psi_\mathrm{c}\) encodes the contact kinematics and constitutive or constraint model, while the components of \(\chi_\mathrm{c}\) collect robot, human-tissue, geometry, and contact-history properties. Together, the equations illustrate how inertia, potential forces, actuation, and feedback evolve the configuration trajectory, which through \(M(q)\) and the contact Jacobians determines reflected inertia, contact geometry, pressure distribution, and \(\tau_\mathrm{c}\). Injury severity, therefore, depends on the coupled contact dynamics between the robot and the human, which are shaped by embodied and motor-cognitive intelligence, human tissue properties, and contact conditions~\cite{haddadin2013towards,haddadin2012making,dickson2025safe,wong2025contact}.

Formal methods provide an established language for specifying and verifying safety properties in autonomous and controlled systems~\cite{luckcuck2019formal}. Recent work translates injury knowledge and contact-force limits into explicit control constraints~\cite{haddadin2012making,dickson2025safe,wong2025contact}, but guarantees remain conditional on the contact and injury models. For a force-based contact-safety requirement, let \(F_\mathrm{c}^{\max}\) denote the context-dependent allowable resultant contact force and, following the sign convention of robustness semantics~\cite{gilpin2021smooth}, define
\begingroup
\small
\begin{equation}
    F_\mathrm{c}(t)=\left\|\int_{\mathcal{A}_\mathrm{c}(t)}n(r,q)p(t,r)\,\mathrm{d}A\right\|,
    \qquad
    \rho_F(t)=1-\frac{F_\mathrm{c}(t)}{F_\mathrm{c}^{\max}}.
    \label{eq:peak_force_safety_metric}
\end{equation}
\endgroup
Here \(F_\mathrm{c}(t)\) is the resultant contact-force magnitude. A positive value of \(\rho_F(t)\) indicates satisfaction of the stated context-specific force limit, zero denotes its boundary, and a negative value indicates a violation. Analogous margins can be defined for pressure, compression, transferred energy, acceleration, or other injury-severity criteria by comparing the relevant quantity with its context-specific limit.

A positive predicted margin may be misleading when perception, human motion, or contact conditions are uncertain. Confidence estimates should be checked against observed accuracy~\cite{guo2017calibration,gawlikowski2023survey}. For action $a$, $L_\alpha(a)$ is the predicted lower $\alpha$-quantile of its minimum safety margin: the actual minimum should fall below this bound in at most an $\alpha$ fraction of validation trials. The robot should select an action $a^*$ with $L_\alpha(a^*)>0$ or invoke a defined fallback.

Design follows from the assessed risks and corresponding margins. Embodied intelligence, through smooth geometry, low inertia, high compliance, and energy dissipation can reduce impact severity even when active collision avoidance fails~\cite{desantis2008atlas,haddadin2013towards}. Active compliance enabled by motor-cognitive intelligence shapes stable, yielding interaction through compliant and energy-aware control across rigid and soft robots~\cite{khan2014compliance,albuschaffer2007unified,dellasantina2023model}. Their guarantees usually concern stability, passivity, or energy~\cite{albuschaffer2007unified} rather than direct satisfaction of an injury threshold. They should therefore form the nominal interaction behavior complemented when necessary by explicit injury-aware or contact-force constraints~\cite{haddadin2012making,dickson2025safe,wong2025contact}.

Motor-cognitive intelligence further extends to active risk control, beginning with human-state estimation and motion prediction to anticipate where and when contact may occur, the relative motion involved, and which body regions are exposed~\cite{rudenko2020human,lasota2017survey}. Confidence-aware and robust predictions can update planning and control constraints~\cite{fisac2018probabilistically}. Their uncertainty should inform conservative margins and task-appropriate clearance, speed~\cite{kirschner2022expectable}, and interaction impedance~\cite{haninger2023model}. Information gathering can defer irreversible actions~\cite{curtis2024partially}. Trials varying noise, occlusion, and conditions outside the training data should test these responses and fallback selection, including whether unnecessary caution subsides as uncertainty resolves.

Safety-aware planning and runtime assurance seek to maintain these bounds through explicit safety constraints and filters~\cite{wabersich2023data}. Safety should shape the nominal policy early: a filter acting only at the boundary may become infeasible or require fallback, intervene abruptly, or sacrifice performance when the planner repeatedly proposes unsafe motion~\cite{hsu2024safety}. These approaches are increasingly being extended across robotic embodiments, including shape and contact-force constraints~\cite{hachen2025non,dickson2025safe,wong2025contact}. Across embodiments, guarantees still depend on model fidelity, observability, real-time computation, and actuator authority~\cite{hsu2024safety}. Robust margins should incorporate uncertainty in human motion and contact parameters, while runtime assurance should specify recovery when these assumptions fail or faults are detected~\cite{laine2020eyes,jing2025fault}.

During contact, tactile sensing and contact localization can support contact detection and appropriate reflex responses~\cite{haddadin2017robot}; learned tactile features can also help distinguish intended physical interaction from unintended collision~\cite{golz2015using}. Communicative intelligence should provide people sharing the workspace with actionable cues about safety-relevant robot context, including intended contact and recovery~\cite{gopinath2019understanding}. These cues should communicate robot intent and status early enough for people to respond~\cite{pascher2023how,habibian2025survey}. Together with speed and separation, they also shape perceived safety~\cite{rubagotti2022perceived}. Their effectiveness should be validated for the intended users and environment. By helping people remain within the robot's actual interaction envelope, they complement mechanical and control safeguards~\cite{iso2010iso12100}.

We therefore propose three linked design commitments: \textit{reduce hazardous contact loads through embodied intelligence}; \textit{maintain safety margins and support recovery through motor-cognitive intelligence}; and \textit{make safety-relevant context actionable through communicative intelligence}. Their effectiveness should be evaluated jointly within the intended interaction context. People must also be able to anticipate what the robot will do and how it will respond to uncertainty. That requirement motivates behavioral intelligibility.

\section*{Behavioral Intelligibility}

Behavioral intelligibility is the system-side capacity comprising \emph{predictability}, whether someone who knows the task and environment can anticipate the robot's motion; and \emph{legibility}, whether observed motion reveals its goal or target~\cite{dragan2013legibility}.
Even if a robot is safe and capable of performing a task, it is not trustworthy if it requires continuous monitoring or moves unpredictably or abruptly.

Failures can occur even when a robot executes its task correctly: an assistive arm may not reveal whether it will hand over or set down a cup; a mobile robot may not signal whether it will yield; and a humanoid’s gaze and reach may indicate different interaction partners or turn-taking roles. Intent may therefore become clear too late for coordination, while hesitation may reflect caution, yielding, replanning, or failure. Conversely, a legible goal does not guarantee predictable motion: abrupt replanning, jitter, or implausible recovery can still violate expectations. In soft robots, distributed deformation and stored elastic energy complicate motion tracking and surface following~\cite{dellasantina2023model}; hysteresis and recoil may cause motion at the contact surface to differ from the commanded motion. 
Reliability drops can reduce real-time trust~\cite{desai2013impact}, while close, rapid approaches may elicit involuntary startle responses~\cite{kirschner2022expectable}. Trust is also shaped by error magnitude~\cite{rossi2017how} and by sequences of errors, particularly near the end of an interaction~\cite{wang2024continuous}. 
Evaluation should therefore account for rare but consequential events: a single unexplained lunge or force spike may outweigh many nominal trials. 

Predictability asks whether a trajectory matches what an observer expects for a known goal, whereas legibility asks whether the observed motion reveals that goal quickly and clearly~\cite{dragan2013legibility}. Let $s$ denote the context, $\mathcal{G}$ the human observer's goal hypotheses, $g^*\in\mathcal{G}$ the robot's actual goal, $\xi^R$ the executed trajectory over $[0,T]$, $\xi^R_{0:N}$ its $N$-step discretization, and $\xi^R_{[0,\tau]}$ the motion observed up to $\tau\in[0,T]$. Let $p_H$ model human expectations and $w(t)$ be a nonnegative weight that decreases over time. We define
\begingroup
\small
\begin{equation}
    \mathcal{M}_{\mathrm{pred}}=\frac{1}{N}\log p_H(\xi^R_{0:N}\mid g^*,s),\qquad
    \ell_{\mathrm{leg}}(\tau)=p_H(g^*\mid\xi^R_{[0,\tau]},s),\qquad
    \mathcal{M}_{\mathrm{leg}}=\frac{\int_0^T w(t)\ell_{\mathrm{leg}}(t)\,dt}{\int_0^T w(t)\,dt}.
    \label{eq:behavioral_intelligibility}
\end{equation}
\endgroup
$\mathcal{M}_{\mathrm{pred}}$ measures predictability as the mean log-likelihood per step of the completed trajectory given the actual goal and context; higher values indicate greater agreement with human expectations. $\ell_{\mathrm{leg}}(\tau)$ measures legibility at time $\tau$ as the probability assigned to the actual goal after observing the motion so far. $\mathcal{M}_{\mathrm{leg}}$ summarizes legibility over the full trajectory, with decreasing weights rewarding earlier goal recognition. Exaggerated motion can therefore increase legibility while reducing predictability~\cite{dragan2013legibility}.
Human demonstrations and interactive feedback can make $p_H$ operational without redefining predictability. Box~\ref{box:demonstration_predictability} connects completed-trajectory likelihood with evidence from corrections and interventions.
As uncertainty increases, motion and fallback behavior should become more predictable, with cues clarifying the robot's next action.
Across methods, evaluation should also report rare or extreme behavior, stress-test occlusions, delays, ambiguous goals, disturbances, and conditions outside the fitting data, and use repeated human studies to assess predictability, surprise, coordination, intervention, and protective withdrawal.

\begin{conceptbox}{Estimating predictability from demonstrations and corrections}{box:demonstration_predictability}
\textbf{Demonstrations as an expectation distribution.} Computational treatments of predictability often assume that the observer and robot share a task cost~\cite{dragan2013legibility}. This is convenient but restrictive: human expectations are rarely available as an explicit cost, and an observer need not mentally reproduce the robot's optimizer. Instead, the motion a human would choose or endorse for a given task and environment state could be collected as a demonstration and treated as evidence of what robot motion that person would find predictable. For a known goal $g^*$ and matched interaction context $s$, let
\[
    \mathcal{D}_{g^*,s}=\{\xi_i^H\}_{i=1}^{N_D},\qquad
    \widehat{p}_H^{\,\mathcal D}(\,\cdot\mid g^*,s)
    :=\operatorname{Fit}\!\left(\mathcal{D}_{g^*,s}\right).
\]
Here, $\operatorname{Fit}$ denotes the selected density or generative model; maximum-entropy inverse reinforcement learning is one option~\cite{ziebart2008maximum}. Substituting $\widehat{p}_H^{\,\mathcal D}$ for $p_H$ in Eq.~\eqref{eq:behavioral_intelligibility} yields the demonstration-based predictability estimate $\widehat{\mathcal{M}}_{\mathrm{pred}}^{\,\mathcal D}$.

\textbf{Interventions and corrections as interactive evidence.} Full demonstrations can be expensive, physically impractical, or mismatched to the robot's embodiment. Interactive robot learning instead provides localized evidence while the robot acts~\cite{celemin2022interactive}. Let $\mathcal{I}_{g^*,s}=\{(\xi^{R,j},z_j,\widetilde{\xi}^{H,j})\}_{j=1}^{N_I}$, where $z_j=1$ denotes an intervention and $\widetilde{\xi}^{H,j}$ an optional corrected motion. A model fitted to $\mathcal{I}_{g^*,s}$ estimates the probability $\widehat q_H^{\,\mathcal I}(Z=0\mid\xi,g^*,s)$ that a trajectory proceeds without intervention. Combining both evidence sources gives
\[
    \widehat p_H^{\,\mathcal D,\mathcal I}(\xi\mid g^*,s)
    \propto
    \underbrace{\widehat p_H^{\,\mathcal D}(\xi\mid g^*,s)}_{
        \substack{\text{likelihood under}\\[-1pt]\text{demonstrations}}}
    \,
    \underbrace{\widehat q_H^{\,\mathcal I}(Z=0\mid\xi,g^*,s)}_{
        \substack{\text{probability of no}\\[-1pt]\text{intervention}}}.
\]
The product reweights the demonstration-derived trajectory distribution toward motions consistent with interactive feedback; proportionality denotes normalization over $\xi$. Corrected motions and uncorrected rollouts can augment the positive evidence used to fit the combined model, while intervened trajectories provide negative evidence for fitting $\widehat q_H^{\,\mathcal I}$. Substituting $\widehat p_H^{\,\mathcal D,\mathcal I}$ for $p_H$ in Eq.~\eqref{eq:behavioral_intelligibility} yields the interaction-refined predictability estimate $\widehat{\mathcal M}_{\mathrm{pred}}^{\,\mathcal D,\mathcal I}$.

Both the demonstration-only and interaction-refined estimates remain proxies whose dependence on the sampled population, context, and feedback protocol should be tested on held-out conditions and evaluated against validated self-report measures of perceived predictability, surprise, and coordination, complemented by behavioral measures such as anticipation accuracy and intervention rate~\cite{hoffman2019evaluating}. 
\end{conceptbox}

\begin{figure}[t!]
    \centering
    \includegraphics[width=\textwidth]{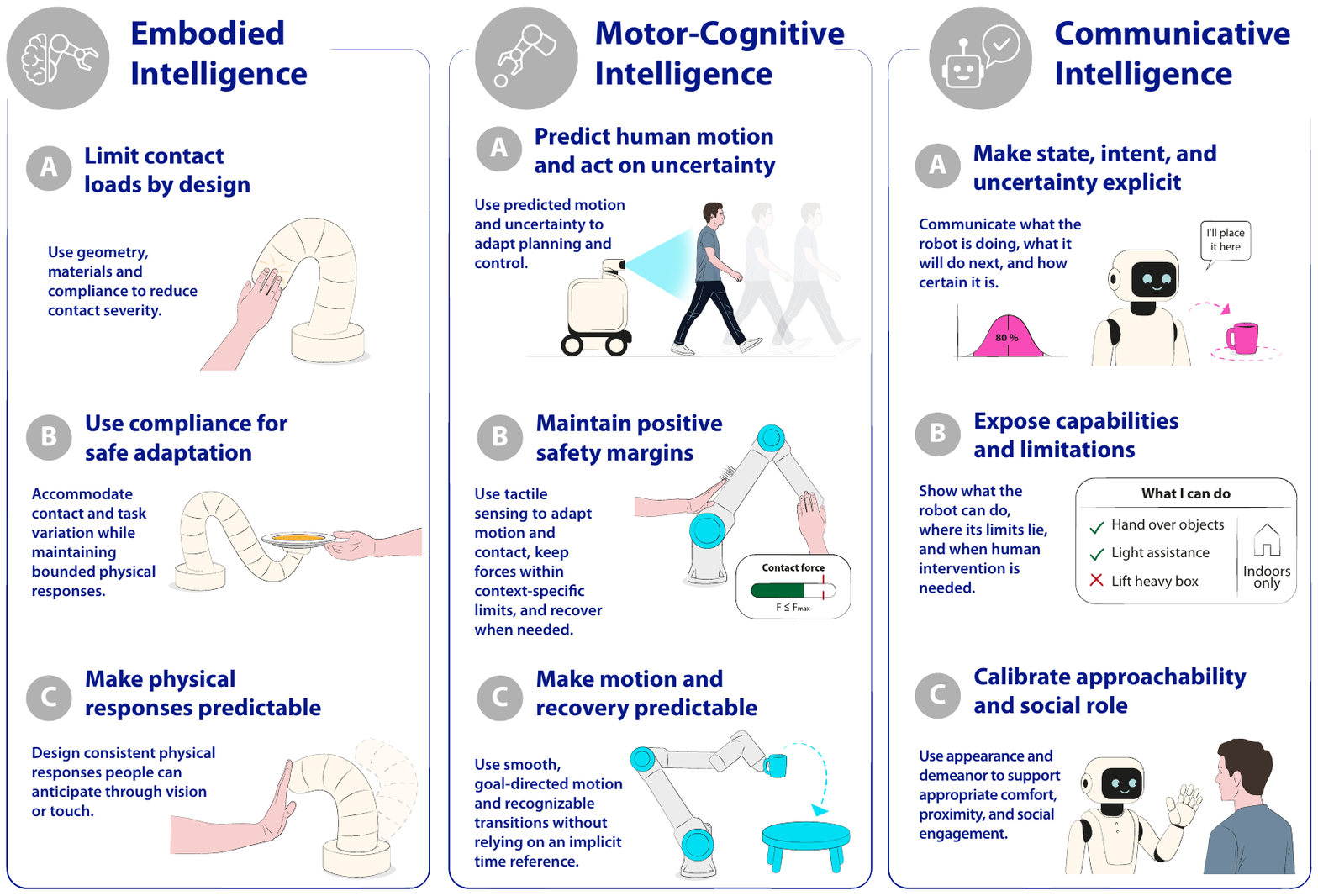}
    \caption{\textbf{Representative design principles for trustworthy physical AI.}
    The three technical pillars provide complementary design levers. \textbf{Embodied Intelligence} supports passive safety through compliant morphology, structural adaptation, and visible and predictable physical responses. \textbf{Motor-Cognitive Intelligence} combines human perception and prediction, uncertainty- and safety-aware control, and predictable motion. \textbf{Communicative Intelligence} communicates robot state, intent, confidence, capabilities, limitations, and social role. Trustworthy interaction requires these principles to be co-designed rather than applied independently.}
    \label{fig:design_principles}
\end{figure}

\Needspace{4\baselineskip}
Robot motion should carry the primary burden of behavioral intelligibility: nominal behavior should be smooth and repeatable, while replanning and recovery should remain recognizable. Designing trajectories that distinguish possible goals from the outset can improve legibility~\cite{dragan2013legibility}, although large deviations from expected motion may reduce predictability~\cite{dragan2013legibility}. When motion remains ambiguous, timely communicative cues can clarify intent and state~\cite{pascher2023how,habibian2025survey}, but should complement rather than compensate for unintelligible behavior~\cite{cini2021relevance}. Learned motion policies should be stress-tested under observation failures and ambiguous human motion to reduce abrupt behavior changes. Outside validated conditions, fallback behavior should be simple and recognizable. When ambiguity cannot be resolved through motion, the robot should explicitly yield or request clarification~\cite{hetherington2021mobile,lidard2024risk}. The goal is a closed loop in which robot motion, human response, and sparse cues progressively clarify intent and upcoming action. We therefore propose three design commitments: make nominal and recovery behavior predictable, make task-relevant goals legible early enough for coordination, and use communication to resolve residual ambiguity rather than compensate for unintelligible motion. Behavioral intelligibility concerns the anticipation of actions and the inference of goals, while perceptual alignment concerns accurate judgments of confidence, capabilities, reliability, and roles.

\section*{Perceptual Alignment}

This section addresses how robots can shape what users perceive them to be, so that perception aligns with what they can actually do. We frame this as the alignment between a robot's actual interaction envelope and the user's inferred model of it, drawing on work in perceived safety, transparency, ergonomics, and calibrated trust~\cite{rubagotti2022perceived,wang2024perceived,lee2004trust}. Misalignment fails in two directions: under-reliance, when capable systems are avoided, and over-reliance, when appearance implies capabilities the robot does not possess, what Brooks calls a violation of the \emph{appearance promise}~\cite{brooks2024three}. Soft robots make this tension especially visible. Their compliant bodies and approachable morphologies are often perceived as gentle, safe, and trustworthy~\cite{dupasquier2025study,wang2024perceived}, yet softness alone guarantees neither bounded forces, robust control, nor transparent failure recovery. Where safety bounds physical risk and behavioral intelligibility makes ongoing action interpretable, Perceptual alignment ensures that users correctly infer a robot's functional capabilities, reliability, compliance, safety limits, and intended role. This calibration has two coupled layers. The first is \emph{epistemic calibration}: whether users accurately infer the robot's capabilities, reliability, uncertainty, and appropriate conditions of use. The second is \emph{social-affective calibration}: whether users adopt an appropriate interpersonal stance toward the robot, including comfort, approachability, and perceived social role. The layers are analytically distinct but practically coupled: the same appearance, motion, softness, or social behavior shapes both what users believe the robot can do and how comfortable they feel interacting with it.

Miscalibration begins before interaction does. People read a robot's appearance and demeanor quickly and largely automatically, and use these cues to infer its abilities, personality, and appropriate role~\cite{goetz2003matching}. In design terms, these are perceived affordances: cues from which users infer what interaction is possible, whether or not the underlying system supports it~\cite{norman2013design}. These attributions move trust independently of capability: the automation literature warned early that humanlike cues invite expectations a system cannot meet~\cite{lee2004trust}, and two decades of meta-analysis show trust in robots increasingly attaching to attributes such as form, personality, and anthropomorphism rather than to measured performance~\cite{hancock2011meta,hancock2021evolving}. The resulting expectation gap fails in both directions~\cite{rosen2022social}. Overestimation produces reliance that observed faults do not correct: participants complied with irreversible and privacy-sensitive requests from a visibly faulty home robot even while rating it less reliable and less trustworthy~\cite{salem2015would}. Underestimation is not mere caution but a self-reinforcing failure: low expectations suppress interaction, so capabilities remain undiscovered, and the inaccurate model persists~\cite{rosen2022social}; a capable manipulator that exposes nothing of its force regulation or active safety invites exactly this failure. The two failures are not symmetric. Overestimation in physical interaction can end in contact the user did not anticipate, while underestimation leaves useful capabilities unused.

The overestimation failure is now being engineered at scale. Home robot products, including several named above, lead with expressive faces, plush or textile surfaces, and curated demonstration videos, cues that build trust through first impressions and media exposure before any capability has been observed~\cite{rosen2022social}. These cues work because their meanings are inherited: decades of fiction have taught audiences to read large eyes and soft features as kindness, and an armored humanoid silhouette as competence. Exposure to fictional robots measurably raises expectations of what real ones can do~\cite{horstmann2019great}, and cuteness raises the perceived trustworthiness of robot faces directly~\cite{chen2023research}, a strategy already described as a dark pattern in home robot design~\cite{lacey2019cuteness}. The two aesthetics divide the labor: companion forms target the social-affective layer of calibration, heroic forms the epistemic one. Little exposure is needed: in a recent caregiving study, a single brief interaction with a soft gripper was enough to significantly increase trust in robots~\cite{dupasquier2025study}. Trust acquired this way is borrowed, not earned. Expectations set high produce sharper disappointment when faults arrive~\cite{paepcke2010judging}, and trust lost to negative events recovers slowly~\cite{lee2004trust}. Whether these design choices survive the first year of deployment is an open and largely untested question.

Evaluating perceptual alignment requires measuring both sides of the alignment. Lee and See provide the constructs for trust in automation: \emph{calibration}, the correspondence between trust and actual capability; \emph{resolution}, whether trust differentiates levels of capability; and \emph{specificity}, whether trust tracks the right function, mode, and moment~\cite{lee2004trust}. A user with poor resolution trusts a robot the same whether it succeeds at grasping 60\% or 90\% of the time; a user with poor specificity extends trust earned in one task or mode to the whole system, assuming a robot that grasps reliably will also hand over safely. These constructs describe the user's judgment, but the burden they imply falls on the robot: it is the system that must afford resolution and specificity, by making differences in capability observable across functions, modes, and time. The field of Human-Robot Interaction has measured the inferred side almost exclusively through single-shot self-report after brief encounters~\cite{hancock2011meta}. This is insufficient for two reasons. First, attitudinal ratings dissociate from behavior~\cite{salem2015would}: reliance, compliance, monitoring, and intervention rates are the more diagnostic signals. Second, calibration is dynamic. First impressions are dominated by appearance priors and novelty effects that decay over weeks~\cite{leite2013social}, trust is history-dependent~\cite{lee2004trust}, and user mental models drift as learned systems drift~\cite{endsley2023supporting}. The alignment target therefore moves on both sides: a learned policy may improve while the user's model stays fixed at what onboarding showed, or hardware may wear while confidence grows through uneventful use, so the overtrusting user after many routine handovers is the case that matters most and the one that single session studies cannot see~\cite{devisser2020towards}. Evaluation should therefore pair perceived and measured interaction envelopes, quantifying how accurately users infer interaction-relevant properties such as confidence, sensing limits, and contact tolerance, favor behavioral over attitudinal evidence, and span repeated interaction rather than a single point~\cite{rosen2022social}.

Designing for calibration starts from matching rather than maximizing. Appearance, demeanor, and social cues should fit a robot's actual role and capability~\cite{goetz2003matching}, and setting expectations low produces less disappointment and more positive appraisal than overselling~\cite{paepcke2010judging}. Calibration requires addressing both directions of misalignment: overtrust can be reduced by revealing the conditions under which the robot may fail, while undertrust can be reduced by demonstrating its capabilities~\cite{devisser2020towards}. Limitations can be communicated through motion that expresses incapability~\cite{kwon2018expressing}, exposure to states where a policy is likely to fail~\cite{huang2018establishing}, and explicit disclosure of partial reliability~\cite{lee2004trust}. Today, this learning is mostly left to chance: drivers of partially automated vehicles discover what the system can and cannot do by watching it fail in the first few drives~\cite{endsley2023supporting}. Capability onboarding would make it deliberate: structured first sessions in which users experience the robot's boundaries before unsupervised use, treating calibration as a prerequisite rather than a byproduct of ownership. Uncertainty, mode changes, and requests for intervention should be salient in interaction-relevant terms, with the caveat that transparency cues must themselves be calibrated: how confidence is presented changes how it is used~\cite{endsley2023supporting}. Current approachable robot designs are converging towards a design language defined by fiction and anime. But it is a vocabulary of warmth and competence, not of capability and limitation, so the cues that would support calibration remain bespoke, reinvented by every platform. The corrective is a deliberate vocabulary: shared, learnable capability and limitation signals, standardized in the way traffic signals are, so calibration transfers across robots rather than restarting with each one.

Perceptual alignment closes the loop of our framework: a robot that is physically safe and behaviorally intelligible still fails as a trustworthy system if users cannot form accurate beliefs about what it can do, when it may fail, and what role it should hold. Engineering this alignment across embodiment, control, and communicative intelligence is the subject of the outlook that follows.

Box~\ref{box:hospital_cup_handover} illustrates how practitioners can specify, evaluate, and co-design these dimensions for a hospital cup handover.

\begin{conceptbox}{Putting the framework into practice:\\ A hospital cup handover with a soft--rigid humanoid}{box:hospital_cup_handover}
Consider a humanoid with soft continuum arms handing a cup of water to an elderly hospital patient.
\begin{enumerate}
\setlength{\topsep}{2pt}
\setlength{\itemsep}{4pt}
\setlength{\parsep}{2pt}
\item \textbf{Identify users, operating conditions, and requirements.}

\textbf{Context:} Identify vulnerable patients, visitors, and medical personnel; specify their abilities, vulnerabilities, expectations, and contact preferences. Define safety limits intended to prevent injury and pain, accounting for each group's proximity, permitted contact, and ability to respond to the robot.
\textbf{Safety:} Identify blunt or sharp contact, gripper pinching, spills, and falls. Set group-specific safety limits and risk tolerance $\alpha$ over a stated horizon. Reliably identify applicable limits; use conservative limits when group membership or vulnerability is uncertain.
\textbf{Intelligibility:} Require predictable arm motion and recognizable handover location, timing, and turn-taking.
\textbf{Perceptual Alignment:} Require a pain-free, comfortable handover, with appearance and interaction cues that accurately convey how the robot limits contact forces and protects the patient.

\item \textbf{Estimate trustworthiness for this interaction.}
Use the figures of merit in Table~\ref{tab:figures_of_merit}.

\textbf{Safety---physical response:} Estimate peak contact loads and minimum safety margins across workspace, speeds, payloads, and actuator limits, accounting for passive structural compliance, controller bandwidth, and stopping or recovery. Include contact along the continuum arms and at the grippers.
\textbf{Safety---uncertainty:} Predict patient and bystander motion; calibrate uncertainty against physical prediction errors and propagate these into safety-margin estimates. Validate risk bounds against $\alpha$ using observed safety-margin violations under representative uncertainties; require a positive conservative margin bound or a defined fallback.
\textbf{Intelligibility:} Collect diverse demonstrations reflecting patient and bystander expectations. Fit an expectation distribution following Box~\ref{box:demonstration_predictability} and evaluate the robot’s policy against it; refine the estimator through corrective feedback, identifying unexpected motion. Evaluate predictability and early recognition of handover goals and timing with held-out participants and interaction conditions.
\textbf{Perceptual Alignment:} Compare the patient's expected contact forces and protective responses with measured behavior, reporting over- and underestimation. Record comfort and withdrawal during the handover.

\item \textbf{Co-design using the measured gaps.}

\textbf{Safety:} Balance arm compliance with cup-support payload capability requirements, add protective padding, shield pinch points around tendons, joints, and grippers, and address balance hazards.
\textbf{Behavioral Intelligibility:} Tune policy rewards to prioritize predictable motion when human-motion uncertainty is high, and task performance, such as completion speed, when uncertainty and proximity permit, while maintaining safety constraints. Preserve smooth transitions between priorities.
\textbf{Perceptual Alignment:} Use discrepancies to adjust physical behavior and interaction cues. Offer optional demonstrations when patients overestimate risk, and adapt the interaction to their comfort and preferences.
\end{enumerate}
Reassess all three dimensions after design changes to identify improvements and tradeoffs.
\end{conceptbox}

\clearpage
\section*{Outlook: Toward a Science of Trustworthy Physical Interaction}

The value of this framework lies in treating trustworthiness as an emergent property of the whole robot. As Figs.~\ref{fig:trust_framework} \&~\ref{fig:design_principles} and Table~\ref{tab:trustworthiness_matrix} illustrate, safety, behavioral intelligibility, and perceptual alignment depend on how embodiment, control, and design work together. A change intended to improve one dimension can alter the others. A more compliant body may soften contact while making motion harder to regulate; greater adaptability may increase capability while reducing predictability; and a more expressive robot may be easier to approach while inviting expectations it cannot meet. Calibrating those expectations to demonstrated capability is a core design responsibility, especially when commercial incentives reward adoption, engagement, and perceived capability. We argue that these couplings should guide design and evaluation jointly. Robots with deformable components (e.g., soft robots~\cite{rus2015design}) provide a particularly revealing setting in which to study these couplings, while the same principles extend to humanoids, collaborative manipulators, assistive devices, domestic robots, and systems governed by end-to-end learned policies. The three dimensions also suggest a research agenda for integrated evaluation, summarized in Table~\ref{tab:figures_of_merit}. Evaluation should characterize trustworthiness as a multidimensional profile tied to explicit users, tasks, and operating conditions, revealing how design choices shift safety, behavioral intelligibility, and perceptual alignment together. Future benchmarks could combine these measures during realistic physical tasks and examine how design choices shift the profile across dimensions. Evaluation should also move from one-shot impressions toward evidence of sustained appropriate reliance. Longitudinal studies will be especially important for understanding how initial impressions develop into stable patterns of reliance over repeated interactions~\cite{leite2013social,lee2004trust}. Digital twins that integrate contact dynamics with models of human expectation and interpretation could further support trustworthiness assessment before deployment.
Over time, insights shared across platforms could also form a recognizable language of physical interaction. Recurring ways of expressing confidence, adaptation, and the need for participation would allow experience with one robot to inform interaction with another. Such a language could connect robot design, evaluation, and standardization around a common understanding of how trustworthy behavior is physically expressed. Trustworthy physical AI should make its capabilities, limits, and uncertainty legible through the combined evidence of its body, behavior, and communication.

\clearpage

\begin{table}
\centering
\caption{\textbf{Figures of merit for evaluation of trustworthy physical interaction.}
The measures operationalize the three system-side dimensions under stated user and operating conditions and should be reported as a multidimensional profile rather than collapsed into a universal trustworthiness score.}
\label{tab:figures_of_merit}
\vspace{0.6em}
\scriptsize
\setlength{\tabcolsep}{3.5pt}
\renewcommand{\arraystretch}{1.16}
\begin{tabularx}{\textwidth}{@{}
  >{\RaggedRight\arraybackslash}p{0.18\textwidth}
  >{\RaggedRight\arraybackslash}p{0.19\textwidth}
  Y@{}}
\toprule
\textbf{Dimension}
&
\textbf{Evaluation target}
&
\textbf{Figure of merit and operationalization}
\\
\midrule

\textbf{Safety}
&
Maintain safety
&
Generalized safety margin $\rho_{\mathrm{safe}}(t)$ (positive satisfies the specified criterion; zero marks its boundary; negative violates it) and trajectory minimum $\rho_{\min}=\min_t\rho_{\mathrm{safe}}(t)$, operationalized via the contact-force margin $\rho_F(t)$ or task-specific separation, pressure, energy, compression, or acceleration margins.
\\
\midrule

\textbf{Safety}
&
Control physical risk under uncertainty
&
$L_\alpha(a)$ is a conservative lower bound on the minimum safety margin predicted across possible outcomes of action $a$; $\alpha$ is the tolerated frequency with which the actual margin may fall below this bound. Report this frequency, the typical gap between the bound and actual margin, and how often the robot selects an action $a$ with $L_\alpha(a)>0$ or invokes a defined fallback. Also report actual violations.
\\
\midrule

\textbf{Safety}
&
Recover from an unsafe state
&
Minimum margin during recovery $\rho_{\min}^{\mathrm{rec}}$ and time $T_{\mathrm{safe}}$ to return to and remain at $\rho_{\mathrm{safe}}>0$; report unrecovered trials separately.
\\
\midrule

\textbf{Behavioral Intelligibility}
&
Produce predictable motion
&
Trajectory predictability $\mathcal M_{\mathrm{pred}}$, including fallback behavior, operationalized via $\widehat{\mathcal M}_{\mathrm{pred}}^{\,\mathcal D}$ or $\widehat{\mathcal M}_{\mathrm{pred}}^{\,\mathcal D,\mathcal I}$; validate against human anticipation accuracy, surprise, intervention, or correction rates as appropriate.
\\
\midrule

\textbf{Behavioral Intelligibility}
&
Reveal the intended goal early
&
Time-weighted legibility $\mathcal M_{\mathrm{leg}}$; report the goal hypothesis set, observation horizon, and time to correct goal inference.
\\
\midrule

\textbf{Perceptual Alignment}
&
Align perceived and actual capability
&
Signed discrepancy between users' elicited estimates of limits relevant to the interaction, such as task success rate, sensing range, and contact tolerance, and the validated values reported in the Safety rows~\cite{okamura2020adaptive,norton2022metrics}. Report overestimation and underestimation separately, since they differ in consequence and remedy. State the user population, prior exposure, and elicitation instrument~\cite{rubagotti2022perceived}. \\ \midrule \textbf{Perceptual Alignment} & Confirm alignment in behavior & Reliance measures collected in the same sessions: intervention and override rate, unnecessary avoidance, and protective withdrawal. Report alongside the elicited estimate, because attitudinal and behavioral measures dissociate~\cite{kohn2021measurement,lee2004trust}. 
\\

\bottomrule
\end{tabularx}
\end{table}

\begin{table}
\centering
\caption{\textbf{Trustworthy robot interaction as a co-design problem.}
Rows represent the system-side dimensions of trustworthiness, i.e., what interaction should achieve; columns represent the technical pillars, or design levers, through which these dimensions can be engineered. Diagonal cells indicate intuitive anchor relationships, whereas off-diagonal cells highlight cross-pillar couplings.}
\label{tab:trustworthiness_matrix}
\scriptsize
\setlength{\tabcolsep}{2.3pt}
\renewcommand{\arraystretch}{1.18}
\begin{tabularx}{\textwidth}{@{}
  >{\RaggedRight\arraybackslash}p{0.18\textwidth}
  YYY@{}}
\toprule
\makecell[l]{\textbf{Technical Pillars} $\rightarrow$\\
\textbf{Trustworthiness}\\
\textbf{Dimensions} $\downarrow$}
&
\textbf{Embodied Intelligence}
&
\textbf{Motor-Cognitive Intelligence}
&
\textbf{Communicative Intelligence}
\\
\midrule

\textbf{Safety}
&
\cellcolor{gray!12}
\textbf{Passive harm mitigation.}
Mechanical compliance, damping, mass distribution, contact area, surface texture, and mechanical stops shape contact severity. Soft or compliant bodies can reduce peak forces, pressures, and transferred energy when perception, planning, or control fail.
&
\textbf{Active risk control.}
Human-aware sensing, human motion prediction, stable control, collision avoidance, safe motion generation, fail-safe recovery, and safety filters actively bound risk under uncertainty in closed-loop interaction. Controller gains, impedance/admittance design, latency, and passivity directly affect safety.
&
\textbf{Behavior-shaping safety cues.}
Mode indicators, motion previews, lights, sounds, displays, transparent warnings, and role-consistent appearance shape human behavior around the robot, including proximity, handover timing, workspace entry, and appropriate reliance.
\\
\midrule

\textbf{Behavioral Intelligibility}
&
\textbf{Intelligible embodied dynamics.}
Passive dynamics, compliant or underactuated morphology, soft contact, and visible shape adaptation create physical regularities that humans can learn. Hidden deformation, hysteresis, nonlinear material response, or oscillation can make behavior harder to predict.
&
\cellcolor{gray!12}
\textbf{Predictable and legible motion.}
Low-level stability and high-level motion policies determine whether actions are smooth, bounded, repeatable, goal-directed, and consistent with human expectations. Perception failures, abrupt replanning, jitter, hidden timing references, or unexplained pauses can make motion appear unpredictable.
&
\textbf{Observable state, intent, and upcoming action.}
Appearance, affordances, gaze-like orientation, body posture, lights, sounds, displays, transparent covers, and motion previews help humans infer the robot's state, goal, mode, turn-taking structure, and upcoming action.
\\
\midrule

\textbf{Perceptual Alignment}
&
\textbf{Physical compliance calibration.}
Visual, tactile, material, geometric, and scale cues should make perceived softness, stiffness, yielding, inertia, and contact tolerance match the robot's actual passive mechanical compliance. This calibrates the embodied component of perceived safety: what the body itself can absorb or limit during contact.
&
\textbf{Competence, reliability, and active-safety calibration.}
Observable motion and autonomy cues, such as smoothness, responsiveness, consistency, recovery, hesitation, interruption, and failure handling, should make perceived confidence, competence, reliability, and active risk control match the robot's safety-aware control capabilities.
&
\cellcolor{gray!12}
\textbf{Aesthetic-social approachability.}
Appearance, morphology, expressiveness, surface-material style, perceived cuteness, and behavioral congruence should make people feel appropriately comfortable approaching, being near, and socially interacting with the robot. These cues should match the robot's intended role and actual social-interaction capabilities.
\\

\bottomrule
\end{tabularx}
\end{table}


\clearpage
\bibliography{main}

@article{kok2020trust,
  title={Trust in robots: Challenges and opportunities},
  author={Kok, Bing Cai and Soh, Harold},
  journal={Current robotics reports},
  volume={1},
  number={4},
  pages={297--309},
  year={2020},
  publisher={Springer}
}

@article{rus2015design,
  title={Design, fabrication and control of soft robots},
  author={Rus, Daniela and Tolley, Michael T},
  journal={Nature},
  volume={521},
  number={7553},
  pages={467--475},
  year={2015},
  publisher={Nature Publishing Group UK London}
}

@article{zlotowski2016appearance,
  title={Appearance of a robot affects the impact of its behaviour on perceived trustworthiness and empathy},
  author={Z{\l}otowski, Jakub and Sumioka, Hidenobu and Nishio, Shuichi and Glas, Dylan F and Bartneck, Christoph and Ishiguro, Hiroshi},
  journal={Paladyn, Journal of Behavioral Robotics},
  volume={7},
  number={1},
  pages={55--66},
  year={2016},
  publisher={De Gruyter Open}
}

@article{he2021challenges,
  title={The challenges and opportunities of human-centered AI for trustworthy robots and autonomous systems},
  author={He, Hongmei and Gray, John and Cangelosi, Angelo and Meng, Qinggang and McGinnity, T Martin and Mehnen, J{\"o}rn},
  journal={IEEE Transactions on Cognitive and Developmental Systems},
  volume={14},
  number={4},
  pages={1398--1412},
  year={2021},
  publisher={IEEE}
}

@article{devisser2020towards,
  title={Towards a theory of longitudinal trust calibration in human--robot teams},
  author={De Visser, Ewart J and Peeters, Marieke MM and Jung, Malte F and Kohn, Spencer and Shaw, Tyler H and Pak, Richard and Neerincx, Mark A},
  journal={International journal of social robotics},
  volume={12},
  number={2},
  pages={459--478},
  year={2020},
  publisher={Springer}
}

@misc{robotreport2025look,
   title = {A look at iRobot’s 35-year robotics journey},
   author = {{The Robot Report}},
   year = {2025},
   url = {https://www.therobotreport.com/a-look-at-irobot-35-year-robotics-journey/}
}

@article{rubagotti2022perceived,
  title={Perceived safety in physical human--robot interaction---A survey},
  author={Rubagotti, Matteo and Tusseyeva, Inara and Baltabayeva, Sara and Summers, Danna and Sandygulova, Anara},
  journal={Robotics and Autonomous Systems},
  volume={151},
  pages={104047},
  year={2022},
  publisher={Elsevier},
  doi={10.1016/j.robot.2022.104047}
}

@article{khan2014compliance,
  title={Compliance Control and Human--Robot Interaction: Part 1---Survey},
  author={Khan, Said G. and Herrmann, Guido and Al Grafi, Mubarak and Pipe, Tony and Melhuish, Chris},
  journal={International Journal of Humanoid Robotics},
  volume={11},
  number={3},
  pages={1430001},
  year={2014},
  publisher={World Scientific},
  doi={10.1142/S0219843614300013}
}

@article{desantis2008atlas,
  title={An atlas of physical human--robot interaction},
  author={{De Santis}, Agostino and Siciliano, Bruno and {De Luca}, Alessandro and Bicchi, Antonio},
  journal={Mechanism and Machine Theory},
  volume={43},
  number={3},
  pages={253--270},
  year={2008},
  publisher={Elsevier},
  doi={10.1016/j.mechmachtheory.2007.03.003}
}

@article{firminodesouza2025trust,
  title={Trust and Trustworthiness from Human-Centered Perspective in Human--Robot Interaction ({HRI})---A Systematic Literature Review},
  author={{Firmino de Souza}, Debora and Sousa, Sonia and Kristjuhan-Ling, Kadri and Dunajeva, Olga and Roosileht, Mare and Pentel, Avar and M{\~o}ttus, Mati and {\"O}zdemir, Mustafa Can and Grat{\v{s}}jova, {\v{Z}}anna},
  journal={Electronics},
  volume={14},
  number={8},
  pages={1557},
  year={2025},
  publisher={MDPI},
  doi={10.3390/electronics14081557}
}

@article{samarathunga2025assessing,
  title={Assessing Safety in Physical Human--Robot Interaction in Industrial Settings: A Systematic Review of Contact Modelling and Impact Measuring Methods},
  author={Samarathunga, S. M. B. P. B. and Valori, Marcello and Legnani, Giovanni and Fassi, Irene},
  journal={Robotics},
  volume={14},
  number={3},
  pages={27},
  year={2025},
  publisher={MDPI},
  doi={10.3390/robotics14030027}
}

@article{mengaldo2022concise,
  title={A concise guide to modelling the physics of embodied intelligence in soft robotics},
  author={Mengaldo, Gianmarco and Renda, Federico and Brunton, Steven L. and B{\"a}cher, Moritz and Calisti, Marcello and Duriez, Christian and Chirikjian, Gregory S. and Laschi, Cecilia},
  journal={Nature Reviews Physics},
  volume={4},
  number={9},
  pages={595--610},
  year={2022},
  publisher={Nature Publishing Group},
  doi={10.1038/s42254-022-00481-z}
}

@inproceedings{dragan2013legibility,
  title={Legibility and predictability of robot motion},
  author={Dragan, Anca D and Lee, Kenton CT and Srinivasa, Siddhartha S},
  booktitle={2013 8th ACM/IEEE International Conference on Human-Robot Interaction (HRI)},
  pages={301--308},
  year={2013},
  organization={IEEE},
  doi={10.1109/HRI.2013.6483603}
}

@article{hoffman2019evaluating,
  title={Evaluating fluency in human--robot collaboration},
  author={Hoffman, Guy},
  journal={IEEE Transactions on Human-Machine Systems},
  volume={49},
  number={3},
  pages={209--218},
  year={2019},
  doi={10.1109/THMS.2019.2904558}
}

@inproceedings{fisac2018probabilistically,
  title={Probabilistically safe robot planning with confidence-based human predictions},
  author={Fisac, Jaime F. and Bajcsy, Andrea and Herbert, Sylvia L. and Fridovich-Keil, David and Wang, Steven and Tomlin, Claire J. and Dragan, Anca D.},
  booktitle={Proceedings of Robotics: Science and Systems},
  year={2018},
  address={Pittsburgh, PA, USA},
  doi={10.15607/RSS.2018.XIV.069}
}

@inproceedings{lidard2024risk,
  title={Risk-calibrated human--robot interaction via set-valued intent prediction},
  author={Lidard, Justin and Pham, Hang and Bachman, Ariel and Boateng, Bryan and Majumdar, Anirudha},
  booktitle={Proceedings of Robotics: Science and Systems},
  year={2024},
  address={Delft, Netherlands},
  doi={10.15607/RSS.2024.XX.027}
}

@inproceedings{hetherington2021mobile,
  title={Mobile Robot Yielding Cues for Human--Robot Spatial Interaction},
  author={Hetherington, Nicholas J. and Lee, Ryan and Haase, Marlene and Croft, Elizabeth A. and Van der Loos, H. F. Machiel},
  booktitle={2021 IEEE/RSJ International Conference on Intelligent Robots and Systems (IROS)},
  pages={3028--3033},
  year={2021},
  organization={IEEE},
  doi={10.1109/IROS51168.2021.9636367}
}

@inproceedings{curtis2024partially,
  title={Partially Observable Task and Motion Planning with Uncertainty and Risk Awareness},
  author={Curtis, Aidan and Matheos, George and Gothoskar, Nishad and Mansinghka, Vikash K. and Tenenbaum, Joshua B. and Lozano-P{\'e}rez, Tom{\'a}s and Kaelbling, Leslie Pack},
  booktitle={Proceedings of Robotics: Science and Systems},
  year={2024},
  address={Delft, Netherlands},
  doi={10.15607/RSS.2024.XX.118}
}

@inproceedings{pascher2023how,
  title={How to communicate robot motion intent: A scoping review},
  author={Pascher, Max and Gruenefeld, Uwe and Schneegass, Stefan and Gerken, Jens},
  booktitle={Proceedings of the 2023 CHI Conference on Human Factors in Computing Systems},
  articleno={562},
  pages={1--17},
  year={2023},
  publisher={Association for Computing Machinery},
  doi={10.1145/3544548.3580857}
}

@article{cini2021relevance,
  title={The relevance of signal timing in human--robot collaborative manipulation},
  author={Cini, Francesca and Banfi, Tommaso and Ciuti, Gastone and Craighero, Laila and Controzzi, Marco},
  journal={Science Robotics},
  volume={6},
  number={58},
  pages={eabg1308},
  year={2021},
  doi={10.1126/scirobotics.abg1308}
}

@article{dupasquier2025study,
  title={A study of perceived safety for soft robotics in caregiving tasks},
  author={{du Pasquier}, Cosima and Grannen, Jennifer and Pan, Chuer and Huber, Serin L and Smith, Aliyah and Kennedy, Monroe and Song, Shuran and Sadigh, Dorsa and Okamura, Allison M},
  journal={arXiv preprint arXiv:2503.20916},
  year={2025}
}

@misc{brooks2024three,
  author       = {Brooks, Rodney A.},
  title        = {Rodney Brooks' Three Laws of Robotics},
  howpublished = {\url{https://rodneybrooks.com/rodney-brooks-three-laws-of-robotics/}},
  year         = {2024},
  note         = {Accessed May 2026}
}

@article{lee2004trust,
  author  = {Lee, John D. and See, Katrina A.},
  title   = {Trust in Automation: Designing for Appropriate Reliance},
  journal = {Human Factors},
  volume  = {46},
  number  = {1},
  pages   = {50--80},
  year    = {2004},
  doi     = {10.1518/hfes.46.1.50_30392}
}

@article{wang2024perceived,
   author = {Yun Wang and Gang Wang and Weihan Ge and Jinxi Duan and Zixin Chen and Li Wen},
   doi = {10.3390/BIOMIMETICS9010058/S1},
   journal = {Biomimetics},
   pages = {58},
   publisher = {Multidisciplinary Digital Publishing Institute (MDPI)},
   title = {Perceived Safety Assessment of Interactive Motions in Human–Soft Robot Interaction},
   volume = {9},
   year = {2024},
}

@book{haddadin2013towards,
  title={Towards safe robots: approaching Asimov’s 1st law},
  author={Haddadin, Sami},
  volume={90},
  year={2013},
  publisher={Springer},
  doi={10.1007/978-3-642-40308-8}
}

@article{haddadin2017robot,
  title={Robot Collisions: A Survey on Detection, Isolation, and Identification},
  author={Haddadin, Sami and {De Luca}, Alessandro and {Albu-Sch\"{a}ffer}, Alin},
  journal={IEEE Transactions on Robotics},
  volume={33},
  number={6},
  pages={1292--1312},
  year={2017},
  doi={10.1109/TRO.2017.2723903}
}

@article{haddadin2012making,
  title={On Making Robots Understand Safety: Embedding Injury Knowledge into Control},
  author={Haddadin, Sami and Haddadin, Simon and Khoury, Augusto and Rokahr, Tim and Parusel, Sven and Burgkart, Rainer and Bicchi, Antonio and {Albu-Sch\"{a}ffer}, Alin},
  journal={The International Journal of Robotics Research},
  volume={31},
  number={13},
  pages={1578--1602},
  year={2012},
  doi={10.1177/0278364912462256}
}

@inproceedings{golz2015using,
  title={Using Tactile Sensation for Learning Contact Knowledge: Discriminate Collision from Physical Interaction},
  author={Golz, Saskia and Osendorfer, Christian and Haddadin, Sami},
  booktitle={2015 IEEE International Conference on Robotics and Automation (ICRA)},
  pages={3788--3794},
  year={2015},
  publisher={IEEE},
  doi={10.1109/ICRA.2015.7139726}
}

@techreport{iso2016ts15066,
  title       = {Robots and robotic devices --- Collaborative robots},
  author      = {{International Organization for Standardization}},
  institution = {International Organization for Standardization},
  address     = {Geneva, Switzerland},
  type        = {Technical Specification},
  number      = {ISO/TS 15066:2016},
  year        = {2016},
  url         = {https://www.iso.org/standard/62996.html}
}

@techreport{iso2010iso12100,
  title       = {Safety of machinery --- General principles for design --- Risk assessment and risk reduction},
  author      = {{International Organization for Standardization}},
  institution = {International Organization for Standardization},
  address     = {Geneva, Switzerland},
  type        = {International Standard},
  number      = {ISO 12100:2010},
  year        = {2010},
  url         = {https://www.iso.org/standard/51528.html}
}

@article{gopinath2019understanding,
  title={Understanding Situational and Mode Awareness for Safe Human--Robot Collaboration: Case Studies on Assembly Applications},
  author={Gopinath, Varun and Johansen, Kerstin},
  journal={Production Engineering},
  volume={13},
  pages={1--9},
  year={2019},
  doi={10.1007/s11740-018-0868-2}
}

@article{celemin2022interactive,
  title={Interactive imitation learning in robotics: A survey},
  author={Celemin, Carlos and P{\'e}rez-Dattari, Rodrigo and Chisari, Eugenio and Franzese, Giovanni and de Souza Rosa, Leandro and Prakash, Ravi and Ajanovi{\'c}, Zlatan and Ferraz, Marta and Valada, Abhinav and Kober, Jens},
  journal={Foundations and Trends{\textregistered} in Robotics},
  volume={10},
  number={1-2},
  pages={1--197},
  year={2022},
  publisher={Emerald Publishing Limited}
}

@inproceedings{ziebart2008maximum,
  title={Maximum entropy inverse reinforcement learning.},
  author={Ziebart, Brian D and Maas, Andrew L and Bagnell, J Andrew and Dey, Anind K and others},
  booktitle={Aaai},
  volume={8},
  pages={1433--1438},
  year={2008},
  organization={Chicago, IL, USA}
}

@article{rudenko2020human,
  title     = {Human motion trajectory prediction: A survey},
  author    = {Rudenko, Andrey and Palmieri, Luigi and Herman, Michael and Kitani, Kris M. and Gavrila, Dariu M. and Arras, Kai O.},
  journal   = {The International Journal of Robotics Research},
  volume    = {39},
  number    = {8},
  pages     = {895--935},
  year      = {2020},
  doi       = {10.1177/0278364920917446}
}

@article{mansfeld2018safety,
  title     = {Safety map: A unified representation for biomechanics impact data and robot instantaneous dynamic properties},
  author    = {Mansfeld, Nico and Hamad, Mazin and Becker, Marvin and Marin, Antonio Gonzales and Haddadin, Sami},
  journal   = {IEEE Robotics and Automation Letters},
  volume    = {3},
  number    = {3},
  pages     = {1880--1887},
  year      = {2018},
  publisher = {IEEE}
}

@article{behrens2022statistical,
  title     = {A statistical model to determine biomechanical limits for physically safe interactions with collaborative robots},
  author    = {Behrens, Roland and Pliske, G and Umbreit, M and Piatek, S and Walcher, F and Elkmann, Norbert},
  journal   = {Frontiers in Robotics and AI},
  volume    = {8},
  pages     = {667818},
  year      = {2022},
  publisher = {Frontiers Media SA}
}

@techreport{iso2023pas5672,
  title       = {Robotics --- Collaborative applications --- Test methods for measuring forces and pressures in human--robot contacts},
  author      = {{International Organization for Standardization}},
  institution = {International Organization for Standardization},
  address     = {Geneva, Switzerland},
  type        = {Publicly Available Specification},
  number      = {ISO/PAS 5672:2023},
  year        = {2023},
  url         = {https://www.iso.org/standard/82488.html}
}

@article{khatib1987unified,
  title     = {A unified approach for motion and force control of robot manipulators: The operational space formulation},
  author    = {Khatib, Oussama},
  journal   = {IEEE Journal on Robotics and Automation},
  volume    = {3},
  number    = {1},
  pages     = {43--53},
  year      = {1987},
  publisher = {IEEE}
}

@article{gawlikowski2023survey,
  title={A survey of uncertainty in deep neural networks},
  author={Gawlikowski, Jakob and Tassi, Cedrique Rovile Njieutcheu and Ali, Mohsin and Lee, Jongseok and Humt, Matthias and Feng, Jianxiang and Kruspe, Anna and Triebel, Rudolph and Jung, Peter and Roscher, Ribana and others},
  journal={Artificial intelligence review},
  volume={56},
  number={Suppl 1},
  pages={1513--1589},
  year={2023},
  publisher={Springer}
}

@article{habibian2025survey,
  title={A survey of communicating robot learning during human-robot interaction},
  author={Habibian, Soheil and Alvarez Valdivia, Antonio and Blumenschein, Laura H and Losey, Dylan P},
  journal={The International Journal of Robotics Research},
  volume={44},
  number={4},
  pages={665--698},
  year={2025},
  publisher={Sage Publications Sage UK: London, England}
}

@inproceedings{guo2017calibration,
  title={On calibration of modern neural networks},
  author={Guo, Chuan and Pleiss, Geoff and Sun, Yu and Weinberger, Kilian Q.},
  booktitle={Proceedings of the 34th International Conference on Machine Learning},
  pages={1321--1330},
  year={2017},
  volume={70},
  series={Proceedings of Machine Learning Research},
  publisher={PMLR}
}

@inproceedings{goetz2003matching,
  author    = {Goetz, Jennifer and Kiesler, Sara and Powers, Aaron},
  title     = {Matching robot appearance and behavior to tasks to improve human-robot cooperation},
  booktitle = {The 12th IEEE International Workshop on Robot and Human Interactive Communication (RO-MAN)},
  year      = {2003},
  pages     = {55--60},
  doi       = {10.1109/ROMAN.2003.1251796}
}

@article{hancock2011meta,
  author  = {Hancock, Peter A. and Billings, Deborah R. and Schaefer, Kristin E. and Chen, Jessie Y. C. and de Visser, Ewart J. and Parasuraman, Raja},
  title   = {A meta-analysis of factors affecting trust in human-robot interaction},
  journal = {Human Factors},
  volume  = {53},
  number  = {5},
  pages   = {517--527},
  year    = {2011},
  doi     = {10.1177/0018720811417254}
}

@article{hancock2021evolving,
  author  = {Hancock, Peter A. and Kessler, Theresa T. and Kaplan, Alexandra D. and Brill, John C. and Szalma, James L.},
  title   = {Evolving trust in robots: Specification through sequential and comparative meta-analyses},
  journal = {Human Factors},
  volume  = {63},
  number  = {7},
  pages   = {1196--1229},
  year    = {2021},
  doi     = {10.1177/0018720820922080}
}

@inproceedings{salem2015would,
  author    = {Salem, Maha and Lakatos, Gabriella and Amirabdollahian, Farshid and Dautenhahn, Kerstin},
  title     = {Would you trust a (faulty) robot? {E}ffects of error, task type and personality on human-robot cooperation and trust},
  booktitle = {Proceedings of the Tenth Annual ACM/IEEE International Conference on Human-Robot Interaction (HRI)},
  year      = {2015},
  pages     = {141--148},
  doi       = {10.1145/2696454.2696497}
}

@inproceedings{desai2013impact,
  author    = {Desai, Munjal and Kaniarasu, Poornima and Medvedev, Mikhail and Steinfeld, Aaron and Yanco, Holly},
  title     = {Impact of Robot Failures and Feedback on Real-Time Trust},
  booktitle = {2013 8th ACM/IEEE International Conference on Human-Robot Interaction (HRI)},
  pages     = {251--258},
  year      = {2013},
  organization = {IEEE},
  doi       = {10.1109/HRI.2013.6483596}
}

@inproceedings{rossi2017how,
  author    = {Rossi, Alessandra and Dautenhahn, Kerstin and Koay, Kheng Lee and Walters, Michael L.},
  title     = {How the Timing and Magnitude of Robot Errors Influence Peoples' Trust of Robots in an Emergency Scenario},
  booktitle = {Social Robotics},
  series    = {Lecture Notes in Computer Science},
  volume    = {10652},
  pages     = {42--52},
  year      = {2017},
  publisher = {Springer International Publishing},
  doi       = {10.1007/978-3-319-70022-9_5}
}

@article{wang2024continuous,
  author  = {Wang, Kexin and Lu, Jianan and Ruan, Shuyi and Qi, Yue},
  title   = {Continuous Error Timing in Automation: The Peak--End Effect on Human--Automation Trust},
  journal = {International Journal of Human--Computer Interaction},
  volume  = {40},
  number  = {8},
  pages   = {1832--1844},
  year    = {2024},
  doi     = {10.1080/10447318.2023.2223954}
}

@article{kirschner2022expectable,
  author  = {Kirschner, Robin Jeanne and Mayer, Henning and Burr, Lisa and Mansfeld, Nico and Abdolshah, Saeed and Haddadin, Sami},
  title   = {Expectable Motion Unit: Avoiding Hazards From Human Involuntary Motions in Human-Robot Interaction},
  journal = {IEEE Robotics and Automation Letters},
  volume  = {7},
  number  = {2},
  pages   = {2993--3000},
  year    = {2022},
  doi     = {10.1109/LRA.2022.3144535}
}

@inproceedings{rosen2022social,
  author    = {Ros{\'e}n, Julia and Lindblom, Jessica and Billing, Erik},
  title     = {The social robot expectation gap evaluation framework},
  booktitle = {Human-Computer Interaction: Technological Innovation},
  series    = {Lecture Notes in Computer Science},
  publisher = {Springer},
  year      = {2022},
  pages     = {590--610},
  doi       = {10.1007/978-3-031-05409-9_43}
}

@article{leite2013social,
  author  = {Leite, Iolanda and Martinho, Carlos and Paiva, Ana},
  title   = {Social robots for long-term interaction: A survey},
  journal = {International Journal of Social Robotics},
  volume  = {5},
  number  = {2},
  pages   = {291--308},
  year    = {2013},
  doi     = {10.1007/s12369-013-0178-y}
}

@article{endsley2023supporting,
  author  = {Endsley, Mica R.},
  title   = {Supporting human-{AI} teams: Transparency, explainability, and situation awareness},
  journal = {Computers in Human Behavior},
  volume  = {140},
  pages   = {107574},
  year    = {2023},
  doi     = {10.1016/j.chb.2022.107574}
}

@inproceedings{kwon2018expressing,
  author    = {Kwon, Minae and Huang, Sandy H. and Dragan, Anca D.},
  title     = {Expressing robot incapability},
  booktitle = {Proceedings of the 2018 ACM/IEEE International Conference on Human-Robot Interaction (HRI)},
  year      = {2018},
  pages     = {87--95},
  doi       = {10.1145/3171221.3171276}
}

@inproceedings{paepcke2010judging,
  author    = {Paepcke, Steffi and Takayama, Leila},
  title     = {Judging a bot by its cover: An experiment on expectation setting for personal robots},
  booktitle = {Proceedings of the 5th ACM/IEEE International Conference on Human-Robot Interaction (HRI)},
  year      = {2010},
  pages     = {45--52},
  doi       = {10.1109/HRI.2010.5453268}
}

@inproceedings{huang2018establishing,
  author    = {Huang, Sandy H. and Bhatia, Kush and Abbeel, Pieter and Dragan, Anca D.},
  title     = {Establishing appropriate trust via critical states},
  booktitle = {IEEE/RSJ International Conference on Intelligent Robots and Systems (IROS)},
  year      = {2018},
  pages     = {3929--3936},
  doi       = {10.1109/IROS.2018.8593649}
}

@article{horstmann2019great,
  author  = {Horstmann, Aike C. and Kr{\"a}mer, Nicole C.},
  title   = {Great expectations? {R}elation of previous experiences with social robots in real life or in the media and expectancies based on qualitative and quantitative assessment},
  journal = {Frontiers in Psychology},
  volume  = {10},
  pages   = {939},
  year    = {2019},
  doi     = {10.3389/fpsyg.2019.00939}
}

@article{chen2023research,
  author  = {Chen, Chien-Hsiung and Jia, Xiaoyu},
  title   = {Research on the influence of the baby schema effect on the cuteness and trustworthiness of social robot faces},
  journal = {International Journal of Advanced Robotic Systems},
  volume  = {20},
  number  = {3},
  year    = {2023},
  doi     = {10.1177/17298806231168486}
}

@inproceedings{lacey2019cuteness,
  author    = {Lacey, Cherie and Caudwell, Catherine},
  title     = {Cuteness as a `dark pattern' in home robots},
  booktitle = {2019 14th ACM/IEEE International Conference on Human-Robot Interaction (HRI)},
  year      = {2019},
  pages     = {374--381},
  doi       = {10.1109/HRI.2019.8673274}
}

@book{norman2013design,
  author    = {Donald A. Norman},
  title     = {The Design of Everyday Things},
  edition   = {Revised and Expanded},
  publisher = {Basic Books},
  address   = {New York},
  year      = {2013}
}

@article{luckcuck2019formal,
	title = {Formal {Specification} and {Verification} of {Autonomous} {Robotic} {Systems}: {A} {Survey}},
	volume = {52},
	issn = {0360-0300},
	shorttitle = {Formal {Specification} and {Verification} of {Autonomous} {Robotic} {Systems}},
	url = {https://dl.acm.org/doi/10.1145/3342355},
	doi = {10.1145/3342355},
	number = {5},
	urldate = {2026-07-13},
	journal = {ACM Computing Surveys (CSUR)},
	author = {Luckcuck, Matt and Farrell, Marie and Dennis, Louise A. and Dixon, Clare and Fisher, Michael},
	month = sep,
	year = {2019},
	pages = {100:1--100:41},
}

@article{belta2019formal,
	title = {Formal {Methods} for {Control} {Synthesis}: {An} {Optimization} {Perspective}},
	volume = {2},
	shorttitle = {Formal {Methods} for {Control} {Synthesis}},
	url = {https://doi.org/10.1146/annurev-control-053018-023717},
	doi = {10.1146/annurev-control-053018-023717},
	number = {1},
	urldate = {2023-06-19},
	journal = {Annual Review of Control, Robotics, and Autonomous Systems},
	author = {Belta, Calin and Sadraddini, Sadra},
	year = {2019},
	note = {\_eprint: https://doi.org/10.1146/annurev-control-053018-023717},
	pages = {115--140},
}

@article{lasota2017survey,
	title = {A {Survey} of {Methods} for {Safe} {Human}-{Robot} {Interaction}},
	volume = {5},
	issn = {1935-8253, 1935-8261},
	url = {https://www.nowpublishers.com/article/Details/ROB-052},
	doi = {10.1561/2300000052},
	language = {English},
	number = {4},
	urldate = {2021-11-09},
	journal = {Foundations and Trends in Robotics},
	publisher = {Now Publishers, Inc.},
	author = {Lasota, Przemyslaw A. and Fong, Terrence and Shah, Julie A.},
	month = may,
	year = {2017},
	pages = {261--349},
}

@article{gilpin2021smooth,
	title = {A {Smooth} {Robustness} {Measure} of {Signal} {Temporal} {Logic} for {Symbolic} {Control}},
	volume = {5},
	issn = {2475-1456},
	url = {https://ieeexplore.ieee.org/document/9114883},
	doi = {10.1109/LCSYS.2020.3001875},
	number = {1},
	urldate = {2025-05-22},
	journal = {IEEE Control Systems Letters},
	author = {Gilpin, Yann and Kurtz, Vince and Lin, Hai},
	month = jan,
	year = {2021},
	pages = {241--246},
}

@article{dickson2025safe,
	title = {Safe {Autonomous} {Environmental} {Contact} for {Soft} {Robots} {Using} {Control} {Barrier} {Functions}},
	volume = {10},
	issn = {2377-3766},
	url = {https://ieeexplore.ieee.org/abstract/document/11160682},
	doi = {10.1109/LRA.2025.3609669},
	number = {11},
	urldate = {2026-05-03},
	journal = {IEEE Robotics and Automation Letters},
	author = {Dickson, Akua K. and Garcia, Juan C. Pacheco and Anderson, Meredith L. and Jing, Ran and Alizadeh-Shabdiz, Sarah and Wang, Audrey X. and DeLorey, Charles and Patterson, Zach J. and Sabelhaus, Andrew P.},
	month = nov,
	year = {2025},
	pages = {11283--11290},
}

@article{wong2025contact,
	title = {Contact-{Aware} {Safety} in {Soft} {Robots} {Using} {High}-{Order} {Control} {Barrier} and {Lyapunov} {Functions}},
	volume = {10},
	issn = {2377-3766},
	url = {https://ieeexplore.ieee.org/abstract/document/11204460},
	doi = {10.1109/LRA.2025.3621965},
	number = {12},
	urldate = {2026-05-05},
	journal = {IEEE Robotics and Automation Letters},
	author = {Wong, Kiwan and Stölzle, Maximilian and Xiao, Wei and {Della Santina}, Cosimo and Rus, Daniela and Zardini, Gioele},
	month = dec,
	year = {2025},
	pages = {12485--12492},
}

@article{albuschaffer2007unified,
	title = {A {Unified} {Passivity}-based {Control} {Framework} for {Position}, {Torque} and {Impedance} {Control} of {Flexible} {Joint} {Robots}},
	volume = {26},
	issn = {0278-3649},
	url = {http://journals.sagepub.com/doi/10.1177/0278364907073776},
	doi = {10.1177/0278364907073776},
	number = {1},
	journal = {The International Journal of Robotics Research},
	author = {Albu-Schäffer, Alin and Ott, Christian and Hirzinger, Gerd},
	month = jan,
	year = {2007},
	pages = {23--39},
}

@article{haninger2023model,
	title = {Model predictive impedance control with {Gaussian} processes for human and environment interaction},
	volume = {165},
	issn = {0921-8890},
	url = {https://www.sciencedirect.com/science/article/pii/S0921889023000702},
	doi = {10.1016/j.robot.2023.104431},
	urldate = {2024-09-30},
	journal = {Robotics and Autonomous Systems},
	author = {Haninger, Kevin and Hegeler, Christian and Peternel, Luka},
	month = jul,
	year = {2023},
	pages = {104431},
}

@article{dellasantina2023model,
	title = {Model-{Based} {Control} of {Soft} {Robots}: {A} {Survey} of the {State} of the {Art} and {Open} {Challenges}},
	volume = {43},
	issn = {1941-000X},
	shorttitle = {Model-{Based} {Control} of {Soft} {Robots}},
	doi = {10.1109/MCS.2023.3253419},
	number = {3},
	journal = {IEEE Control Systems Magazine},
	author = {Della Santina, Cosimo and Duriez, Christian and Rus, Daniela},
	month = jun,
	year = {2023},
	note = {Conference Name: IEEE Control Systems Magazine},
	pages = {30--65},
}

@article{hachen2025non,
  author={Hachen, Maximilian and Shentu, Chengnan and Lilge, Sven and Burgner-Kahrs, Jessica},
  journal={IEEE Robotics and Automation Letters}, 
  title={A Non-Linear Model Predictive Task-Space Controller Satisfying Shape Constraints for Tendon-Driven Continuum Robots}, 
  year={2025},
  volume={10},
  number={3},
  pages={2438-2445},
  doi={10.1109/LRA.2025.3532159}}

@article{jing2025fault,
	title = {Fault {Detection} and {Response} for {Safe} {Control} of {Artificial} {Muscles} in {Soft} {Robots}},
	volume = {9},
	issn = {2475-1456},
	url = {https://ieeexplore.ieee.org/abstract/document/11165112},
	doi = {10.1109/LCSYS.2025.3610637},
	urldate = {2026-05-12},
	journal = {IEEE Control Systems Letters},
	author = {Jing, Ran and Hook, Charles Van and Yang, Ilyoung and Sabelhaus, Andrew P.},
	year = {2025},
	pages = {2321--2326},
}

@article{wabersich2023data,
	title = {Data-{Driven} {Safety} {Filters}: {Hamilton}-{Jacobi} {Reachability}, {Control} {Barrier} {Functions}, and {Predictive} {Methods} for {Uncertain} {Systems}},
	volume = {43},
	issn = {1941-000X},
	shorttitle = {Data-{Driven} {Safety} {Filters}},
	url = {https://ieeexplore.ieee.org/abstract/document/10266799},
	doi = {10.1109/MCS.2023.3291885},
	number = {5},
	urldate = {2026-07-15},
	journal = {IEEE Control Systems Magazine},
	author = {Wabersich, Kim P. and Taylor, Andrew J. and Choi, Jason J. and Sreenath, Koushil and Tomlin, Claire J. and Ames, Aaron D. and Zeilinger, Melanie N.},
	month = oct,
	year = {2023},
	pages = {137--177},
}

@inproceedings{laine2020eyes,
  title     = {Eyes-Closed Safety Kernels: Safety of Autonomous Systems Under Loss of Observability},
  author    = {Laine, Forrest and Chiu, Chih-Yuan and Tomlin, Claire},
  booktitle = {Proceedings of Robotics: Science and Systems},
  address   = {Corvallis, OR, USA},
  year      = {2020},
  doi       = {10.15607/RSS.2020.XVI.096}
}

@article{bajcsy2021robust,
  title   = {A Robust Control Framework for Human Motion Prediction},
  author  = {Bajcsy, Andrea and Bansal, Somil and Ratner, Ellis and Tomlin, Claire J. and Dragan, Anca D.},
  journal = {IEEE Robotics and Automation Letters},
  volume  = {6},
  number  = {1},
  pages   = {24--31},
  year    = {2021},
  doi     = {10.1109/LRA.2020.3028049}
}

@article{hsu2024safety,
  title   = {The Safety Filter: A Unified View of Safety-Critical Control in Autonomous Systems},
  author  = {Hsu, Kai-Chieh and Hu, Haimin and Fisac, Jaime F.},
  journal = {Annual Review of Control, Robotics, and Autonomous Systems},
  volume  = {7},
  pages   = {47--72},
  year    = {2024},
  doi     = {10.1146/annurev-control-071723-102940}
}

@article{okamura2020adaptive, author = {Okamura, Kazuo and Yamada, Seiji}, title = {Adaptive trust calibration for human-AI collaboration}, journal = {PLOS ONE}, volume = {15}, number = {2}, pages = {e0229132}, year = {2020}, doi = {10.1371/journal.pone.0229132}}

@article{kohn2021measurement, author = {Kohn, Spencer C. and de Visser, Ewart J. and Wiese, Eva and Lee, Yi-Ching and Shaw, Tyler H.}, title = {Measurement of Trust in Automation: A Narrative Review and Reference Guide}, journal = {Frontiers in Psychology}, volume = {12}, pages = {604977}, year = {2021}, doi = {10.3389/fpsyg.2021.604977}}

@article{norton2022metrics, author = {Norton, Adam and Admoni, Henny and Crandall, Jacob and Fitzgerald, Tesca and Gautam, Alvika and Goodrich, Michael and Saretsky, Amy and Scheutz, Matthias and Simmons, Reid and Steinfeld, Aaron and Yanco, Holly}, title = {Metrics for Robot Proficiency Self-assessment and Communication of Proficiency in Human-robot Teams}, journal = {ACM Transactions on Human-Robot Interaction}, volume = {11}, number = {3}, pages = {1--38}, year = {2022}, doi = {10.1145/3522579}}
\bibliographystyle{sciencemag}


\section*{Acknowledgments}
\paragraph*{Funding:}
The work by M.~St\"olzle and D.~Rus was supported in part by the Singapore MIT Alliance on Research and Technology (SMART) under the Mens, Manus, et Machina (M3S) program, by the BARI EMERGE program under grant N00014-26-1-2304, and by the MIT-GIST collaboration.
The work by A.~Sabelhaus was supported in part by U.S. National Science Foundation Awards 2340111 and 2432394. 

\end{document}